# Phonikud: Overcoming Phonetic Underspecification for Hebrew Text-To-Speech

*Yakov Kolani*[1], *Maxim Melichov*[2], *Cobi Calev*[3], *Morris Alper*[4,5]

[1] Independent Researcher [2] Reichman University, Israel [3] Cisco Systems [4] Tel Aviv University, Israel
[5] Carnegie Mellon University, USA

## Abstract

Text-to-speech (TTS) for Modern Hebrew is challenged by the language's orthographic complexity, with existing solutions ignoring underspecified phonetic features such as stress. We present a framework for more phonetically accurate Hebrew TTS with four contributions: (1) *Phonikud*, an open-source Hebrew grapheme-to-phoneme (G2P) system that outputs fully-specified International Phonetic Alphabet (IPA) transcriptions, designed by augmenting a base diacritizer. (2) The *ILSpeech* corpus of paired Hebrew audio, text, and expert IPA annotations. (3) A benchmark for the previously unmeasured task of Hebrew G2P conversion. (4) Hebrew audio-to-IPA models capturing previously disregarded phonetic details for automatic TTS evaluation. Our results show that Phonikud more accurately predicts Hebrew phonemes than prior methods, and that small, local TTS models with phonetic input from Phonikud approach large proprietary systems. We release our code, data, and models at `https://phonikud.github.io`.

## 1. Introduction

Despite the Modern Hebrew language being spoken by approximately nine million people [1], it currently lacks an open-source text-to-speech (TTS) system with adequate performance, crucial for applications such as screen readers and smart home technology. Standard techniques perform poorly due to *phonetic underspecification* in the Hebrew script, which omits key phonetic details for the reader to infer from context. For instance, the word ספר may be read as /sˈefer/[1] ("book"), /sapˈar/ ("barber"), /safˈar/ ("he counted"), or /sfar/ ("suburb"). Optional vowel diacritics (*nikud*) are mostly confined to pedagogical texts such as dictionaries and still leave phonetic ambiguity. For example, the vocalized (with vowel diacritics) word בִּירָה may be read as either /bˈira/ ("beer") or /birˈa/ ("capital city"), as the vowel diacritics do not specify lexical stress. This challenges TTS systems, which must receive unvocalized Hebrew text as input.

Existing approaches either map unvocalized Hebrew text directly to audio [2, 3] or predict vowel diacritics and train TTS models on vocalized text [4, 5]. However, these approaches fail to fully resolve phonetic underspecification as in the example above, leading to salient phonetic inaccuracies. Moreover, existing automatic evaluation methods cannot detect such failures. The standard approach is to apply automatic speech recognition (ASR) to synthesized audio and compare to the input text, but this is ineffective for Hebrew: standard Hebrew ASR models output unvocalized text, making them blind to errors in vowel quality and stress placement.

We address both the production and evaluation gaps with a unified framework, consisting of four contributions: (1) We introduce *Phonikud*, a grapheme-to-phoneme (G2P) conversion pipeline that resolves phonetic ambiguities in written Hebrew. We adapt an existing state-of-the-art model for predicting Hebrew vowel diacritics [6], adding lightweight adaptors to predict additional phonetic features needed for unambiguous G2P conversion. This enables Hebrew TTS models trained on a fully-specified phonemic representation. (2) We contribute the *ILSpeech* corpus, consisting of high-quality Hebrew speech with paired audio, text, and expert International Phonetic Alphabet (IPA) annotations. (3) Using ILSpeech, we establish a benchmark for Hebrew G2P, capturing previously unmeasured features such as stress placement. (4) We train Hebrew audio-to-IPA ASR models on ILSpeech, enabling automatic TTS evaluation that captures phonetic details that were previously disregarded. Our experiments show that these contributions, taken together, enable small, local Hebrew TTS models (<100M parameters) to approach the quality of large proprietary systems, confirmed by both automatic metrics and user preferences.

We release[2] our data, code, and models to support the development of accessible Hebrew speech technology.

## 2. Phonetic Underspecification in Hebrew

Hebrew is normally written without vowel marks (*unvocalized text*), but even when these are added (*vocalized text*) it is still underspecified for various phonetic features that are needed for accurate TTS. These may be split into three primary issues:

**Stress.** Lexical stress is only partially predictable from word shape and part of speech in Hebrew [7]. As illustrated by the minimal pair /txˈina/ ("tahini") vs. /txinˈa/ ("grinding"), both spelled טְחִינָה, stress is not indicated in the orthography even when vowel marks are provided.

**Shva.** The vowel mark *shva* is polyvalent, being either silent or pronounced as /e/ depending on complex morpho-phonological rules with many irregularities [8]. For example, in בְּלוֹנְדוֹן /belˈondon/ ("in London") the shva vowel between the first two consonants is pronounced, while in בְּלוֹנְדִּינִי /blondˈini/ ("blonde") it is silent.

**Irregular words.** Infrequently, words may deviate from regular pronunciation rules. A notable case is loanwords containing the phoneme /w/, written identically to /v/. For example, פִּינְגְּוִין /pˈingwin/ ("penguin") is indistinguishable from the hypothetical form */pˈingvin/.

[1] Following TTS conventions, we place stress marks directly before the stressed vowel (/sˈefer/) rather than the syllable onset (/ˈsefer/).

[2] `https://phonikud.github.io`

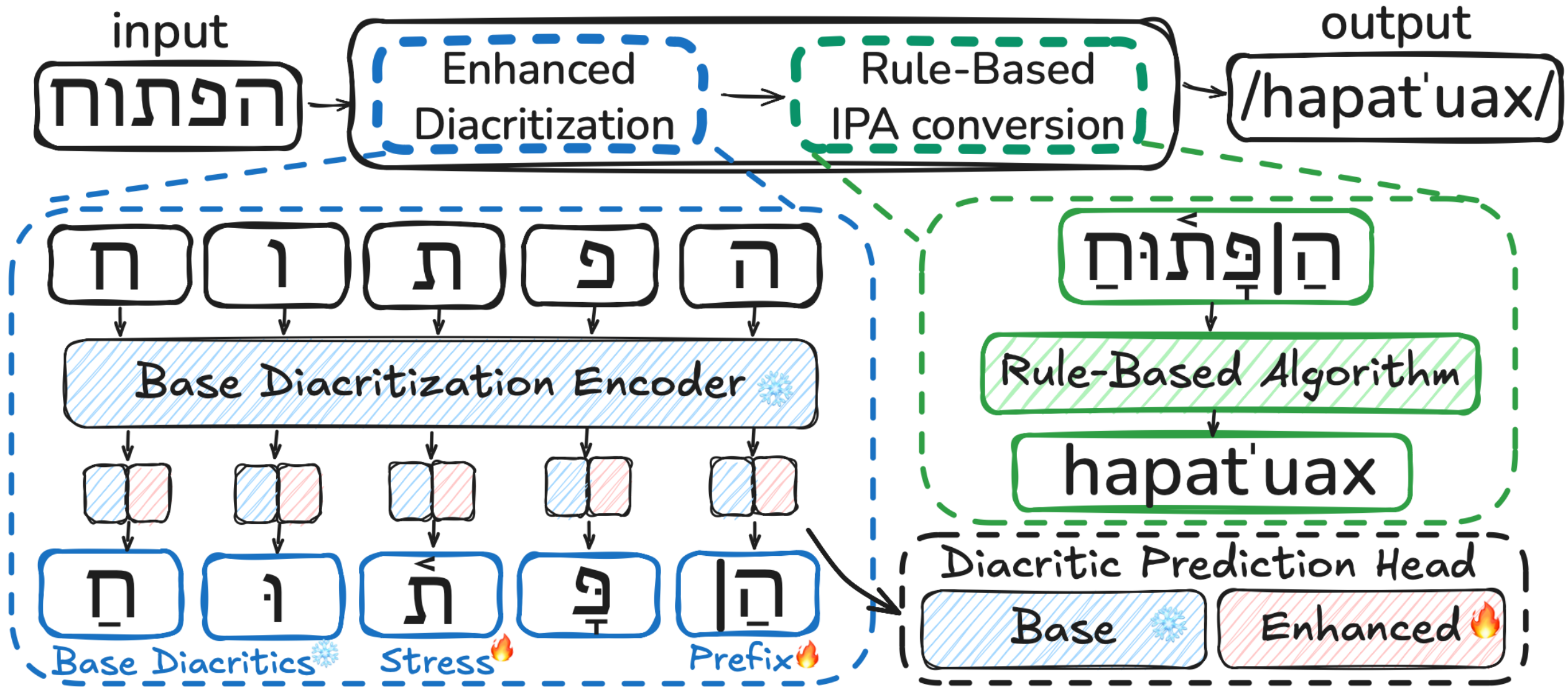

Figure 1: ***The Phonikud grapheme-to-phoneme pipeline.*** *Phonikud converts unvocalized Hebrew text into fully-specified IPA in two steps: First, an enhanced diacritization module adds standard vowel marks and enhanced phonetic symbols to each letter. This uses a frozen (ice symbol) base diacritization model and its per-character prediction head, augmented with a trainable (fire symbol) linear adaptor for predicting enhanced diacritics that disambiguate phonetic ambiguities. Second, a rule-based transformation module converts this text into IPA, which may be used to train performant TTS models.*

## 3. Method

Our Phonikud system is illustrated in Figure 1. As a G2P pipeline, Phonikud takes unvocalized Hebrew text as input and outputs fully-specified IPA transcriptions, which can then be used to train efficient Hebrew TTS systems. We proceed to describe the two key components of this system—an enhanced diacritization module (Section 3.1) and rule-based IPA conversion module (Section 3.2)—followed by the novel procedure used for training the system's learnable components (Section 3.3).

### 3.1. Enhanced Diacritization

Rather than learning to transcribe from scratch, we leverage existing models that efficiently add vowel diacritics to Hebrew text with high accuracy. Our key insight is to augment these with lightweight prediction heads for additional symbols that are needed to resolve phonetic ambiguity. This adds a negligible number of parameters, enabling efficient inference and keeping performance on standard diacritic prediction fixed.

As our base diacritization model, we adopt DictaBERT [6] fine-tuned for Hebrew diacritization (`dicta-il/dictabert-large-char-menaked`), a ∼300M parameter encoder with linear token classification heads for Hebrew diacritics. We freeze this model and add three enhanced diacritics that can be predicted via a new, trainable two-layer multilayer perceptron head (hidden dimension 256, ReLU activation). These new symbols are: (1) a superscript angle indicating a non-final stressed syllable (e.g. לֶ֫חֶם), (2) a subscript line indicating a shva vowel pronounced as /e/ (e.g. מְֽתִיחָה), and (3) a vertical bar indicating the end of a cliticized prefix (e.g. הַ|קוֹד). The first two directly indicate missing phonetic features, while the last aids dictionary matching of irregular words, resolving the issues from Section 2. These graphemes, which include traditional Biblical cantillation marks, are chosen because they are not used in ordinary writing.

### 3.2. Rule-Based IPA Conversion

After generating enhanced vocalized forms (e.g. לֶ֫חֶם), the phonemic representation can be unambiguously determined. We apply a deterministic, rule-based algorithm to convert this to standard IPA (e.g. /ˈlexem/), primarily implemented with a finite-state transducer and dictionary matching. This handles orthographic complexities such as many-to-one grapheme mappings (e.g. ט and ת both represent /t/), non-monotonic sequences (e.g. רֵיחַ represents /rˈeax/ and not the linear order reading */rˈexa/), dual-function letters (e.g. ו may function as a consonant or vowel), and irregular words.

### 3.3. Training Procedure

A fundamental challenge is the lack of existing ground-truth (GT) annotations for Hebrew phonetic features like lexical stress. To address this limitation, we employ a human-in-the-loop procedure to distill knowledge from existing resources along with manual refinement. We semi-automatically annotate a large-scale Hebrew corpus with stress placement, prefix boundaries, and shva realization, and then distill this knowledge into our model via fine-tuning.

To produce large-scale data with these pseudo-GT annotations, we adopt the IsraParlTweet corpus consisting of 5M lines of Hebrew text [9]. We leverage Dicta's (`https://dicta.org.il`) morpho-phonological analysis API and known linguistic rules to automatically predict stress placement, prefix boundaries, and shva realization. Since this is often inaccurate, we manually correct errors by sorting word types (after clitic segmentation) by frequency, correcting a set including the 1K most common items. We train our G2P model on these pseudo-labels until early stopping is triggered (∼6 epochs, batch size 256, learning rate 5e-3, 5% validation split).

# 4. ILSpeech

## 4.1. Corpus Description

We introduce *ILSpeech*, a high-quality Hebrew speech corpus with paired audio, text, and expert-annotated phonetic transcriptions. As existing Hebrew audio corpora [10, 11, 12, 13, 4, 14, 15] lack IPA transcriptions, ILSpeech newly enables evaluation of previously unmeasured features such as stress placement. ILSpeech serves two complementary roles: (1) its text and IPA annotations are used to benchmark G2P conversion (Section 5.1), and (2) its audio-IPA pairs supervise audio-to-IPA ASR models (Section 4.2) for automatic TTS evaluation, capturing phonetic details disregarded by standard Hebrew ASR.

ILSpeech consists of approximately two hours of Hebrew speech from two speakers. The speech content includes diverse topics covering science, technology, history, and everyday conversational speech. Audio was recorded in studio at 44kHz, enhanced using Adobe Enhance [16], and normalized to 22.05kHz. It was then segmented with voice activity detection with manual refinement, yielding 4–14s segments. Hebrew and IPA transcripts were added via expert annotation. Explicit written consent was obtained from the speakers, who were informed about the intended use and data licensing. We release ILSpeech under a non-commercial license with ethical use requirements.

## 4.2. ASR for Hebrew Audio-to-IPA

As discussed in Section 1, standard Hebrew ASR outputs unvocalized text, making it blind to phonetic features such as vowel quality and stress. This limits its applicability to automatic evaluation of TTS systems, where ASR is applied to generations and compared to the original input text. To overcome this issue, we use ILSpeech to train ASR models for Hebrew audio-to-IPA conversion, enabling automatic TTS evaluation that captures these phonetic details. For our tests, we fine-tune a Hebrew ASR model (`openai/whisper-small`) on IPA-audio pairs from ILSpeech. When evaluated on a held-out subset of 150 synthetic utterances with manually-verified IPA transcriptions, this achieves a word error rate (WER) of 24.71% and a character error rate (CER) of 5.47%, sufficient for our purposes (supported by our complementary TTS user study; see Section 5.2). We use this model for automatic TTS metrics as described below, and release it for community use in benchmarking future Hebrew TTS.

# 5. Experiments

We evaluate G2P conversion with Phonikud against existing baselines (Section 5.1), compare downstream TTS performance (Section 5.2), and ablate key system components (Section 5.3). All training runs were conducted on a single RTX4090 GPU (24GB VRAM). Reported error rate metrics are aggregated at the sample level, and p-values are calculated using paired Wilcoxon signed-rank tests (with Bonferroni correction, when multiple tests are performed). Full qualitative results are provided at our project page: `https://phonikud.github.io`.

## 5.1. G2P Evaluation

Phonikud G2P results are shown in Table 1, benchmarked against GT IPA annotations from ILSpeech. Due to compute requirements of some models in our comparison, we evaluate on a random subset of 100 samples; our statistical tests below confirm significance at this sample size. We calculate word- and character error rates (WER, CER), WER ignoring stress ($WER^\sigma$), and exact match (EM; entire output is correct) percentages. We compare to two baselines: the existing state-of-the-art Hebrew diacritizers DictaBERT [6] and Nakdimon [17] with our IPA conversion (using reasonable defaults for ambiguous features such as stress), and multilingual G2P libraries ostensibly supporting Hebrew—eSpeak NG [18] and CharsiuG2P [19].

Table 1: ***G2P evaluation*** *benchmarked on ILSpeech (random 100-item subset). Best results are in* **bold** *(separately for realtime and LLM-based systems). All values are percentages. We illustrate performance on a Hebrew phrase with GT /*bˈoker tˈov*/.* [*] *Diacritizers use our IPA conversion with defaults for ambiguous features like stress.*

| Model | WER↓ | $WER^\sigma$↓ | CER↓ | EM↑ | בוקר טוב |
|---|---|---|---|---|---|
| *Realtime Systems* | | | | | |
| **Phonikud (Ours)** | **17.4** | **12.2** | **3.8** | **17.0** | bˈoker tˈov |
| **Diacritizers*** | | | | | |
| DictaBERT | 39.5 | 25.6 | 8.3 | 2.0 | bokˈer tˈov |
| Nakdimon | 40.5 | 27.2 | 8.7 | 2.0 | bokˈer tˈov |
| **Multilingual G2P** | | | | | |
| eSpeak NG | 100.0 | 95.1 | 44.0 | 0.0 | vvkr tov |
| CharsiuG2P | 100.0 | 100.0 | 70.6 | 0.0 | boːʔab têːb |
| *LLM-based Systems (non-realtime)* | | | | | |
| Claude Opus 4.6 | 25.3 | 19.0 | 5.1 | 10.0 | bˈoker tˈov |
| Gemini 3.1 Pro | **13.9** | **11.8** | **2.8** | **16.0** | bˈoker tˈov |

We also provide G2P results using large language model (LLM) prompting with two leading models: Claude Opus 4.6 [20] and Gemini 3.1 Pro [21], both using chain-of-thought reasoning (high thinking) and applied to individual sentences. While these large, proprietary models are costly and too slow for realtime applications, they are given as a potential upper bound on performance. Inference used temperature 1.0 and a fixed prompt specifying IPA conventions, including stress mark positioning and punctuation retention.

Our system outperforms all realtime baselines and approaches the best LLM-based system. Stress prediction significantly improves performance over existing diacritizers, while multilingual G2P systems are nearly unusable for Hebrew due to limited dictionary support for common words (eSpeak NG) and extensive hallucinations (CharsiuG2P). This advantage is confirmed as statistically significant (all comparisons to Phonikud have $p < 10^{-10}$). Errors in our method's outputs stem from both occasional mistakes in predicting features such as stress, as well from limitations of the base diacritization model.

## 5.2. Downstream TTS Evaluation

Evaluations and comparisons of downstream TTS performance are shown in Tables 2 and 3. We fine-tune the light-weight Piper (VITS) [22, 23] and StyleTTS2 [24] architectures on synthetic Hebrew audio (20 hours, via Gemini 2.5 Pro)—converting GT undiacritized Hebrew transcriptions to IPA with Phonikud. We initialize from pre-trained English TTS checkpoints and train for $\sim$10 hours using default hyperparameters. For Piper, we use the High model size (32M parameters). During inference, we use Phonikud to convert input Hebrew text to IPA used for synthesis. We compare to various open-source Hebrew and multilingual TTS models: MMS [5], the SASPEECH baseline [4], and Robo-

Table 2: ***TTS comparison.*** *Word and character error rates (WER, CER), real-time factor (RTF), and parameter counts for leading Hebrew TTS models, tested on SASPEECH (100 samples). Best results are in **bold** (separately for open and proprietary systems). Error rates are percentages.*

| Model | WER↓ | CER↓ | RTF↓ | # Params |
|---|---|---|---|---|
| *Open Systems* | | | | |
| **Phonikud-Based Models** | | | | |
| Piper (High) | 44.3 | 10.5 | **0.13** | 32M |
| StyleTTS2 | **35.2** | **8.9** | 0.50 | 90M |
| **Open Source Models** | | | | |
| Robo-Shaul | 50.4 | 14.9 | 1.58 | 28M |
| SASPEECH | 68.1 | 21.0 | 0.16 | 28M |
| MMS | 63.6 | 19.6 | 0.21 | 36M |
| HebTTS | 68.7 | 23.3 | 25.4 | 420M |
| *Proprietary Systems* | | | | |
| Gemini | **29.4** | **6.6** | **0.80** | — |
| OpenAI | 35.0 | 8.7 | 1.60 | — |

Table 3: ***TTS stress evaluation.*** *We manually evaluate the frequency of incorrectly synthesized stress on 250 Hebrew sentences with challenging lexical stress patterns, comparing our full method, our method without stress (ours$-\sigma$), and a leading prior model. We report word error rate (WER) and exact match frequency (EM; all words in sentence have correct stress), with best results given in **bold**. All values are percentages.*

| | WER↓ | EM↑ |
|---|---|---|
| Ours | **3.2** | **77.0** |
| Ours$-\sigma$ | 8.4 | 46.4 |
| Robo-Shaul | 6.6 | 56.8 |

Shaul [25] all use two-stage approaches (diacritization followed by speech synthesis), while HebTTS [2] synthesizes speech end-to-end from undiacritized Hebrew text. We also compare to proprietary systems (Google's Gemini 2.5 Flash [26], OpenAI's GPT-4o mini [27]). The evaluation sentences from SASPEECH are in-distribution for some of these models (unlike our method).

Table 2 reports error rates calculated using our audio-to-IPA ASR (Section 4.2), along with Real-Time Factor (RTF) and parameter counts. Open-source models are evaluated on standard consumer hardware (macOS M1) without GPU acceleration to reflect edge computing scenarios, while proprietary models use their cloud APIs. We evaluate on 100 randomly selected SASPEECH [4] samples containing at least six words without special characters. We achieve superior phonetic accuracy and a favorable trade-off between latency and performance compared to open models, confirmed as statistically significant (all comparisons to StyleTTS2 with Phonikud have $p < 10^{-8}$). We also achieve competitive results relative to proprietary systems.

We further test downstream quality with a comparative mean opinion score (CMOS) user study (9 native Hebrew speakers, 20 sentences covering diverse everyday topics) using a 7-point scale (-3 to +3), comparing our strongest system (StyleTTS2 with Phonikud) to the leading open system (Robo-Shaul). Our system is preferred with CMOS +1.3 for naturalness and +0.7 for content fidelity (naturalness $p < 10^{-4}$, content fidelity $p < 0.01$), validating the superior practical utility of TTS trained with Phonikud.

Table 4: ***Ablation study*** *showing downstream effects on TTS. Best results are given in **bold**, and all values are percentages. Removing our enhanced diacritics and basic vowels both degrade performance, while IPA conversion has little effect.*

| Model | WER↓ | CER↓ |
|---|---|---|
| Ours | **49.9** | **13.3** |
| -enhanced diacritization | 55.0 | 15.5 |
| -IPA conversion | 56.1 | 15.4 |
| -vowel diacritics | 69.4 | 23.2 |

To quantify our method's effect on phonetic details salient to native speakers, we curate a test set of 250 sentences covering Hebrew lexical stress ambiguities (i.e., homographs with context-dependent stress) and manually evaluate stress accuracy in generations (Table 3). We compare our full method, our method without stress (trained and evaluated without IPA stress marks), and a leading baseline (Robo-Shaul). Our full method performs significantly better at stress placement ($p < 10^{-7}$ for all comparisons), confirming that other methods more frequently create perceptually salient stress errors.

### 5.3. Ablations

We ablate key parts of our system in Table 4, testing their effect on downstream TTS performance. For these tests, we fix the base TTS model and training settings – using Piper Medium (20M parameters) and using fewer training iterations and less data than in the comparisons above, for a light-weight ablation study. All comparisons between our full system and other settings are statistically significant (all $p < 0.004$). Removing enhanced diacritization hurts performance overall, consistent with our manual evaluation above. Removing IPA conversion (i.e. training directly on vocalized Hebrew text) has a minimal effect, showing that IPA conversion preserves the essential phonetic content needed to synthesize speech, while providing practical advantages such as interpretability and cross-lingual standardization. Training directly on undiacritized Hebrew text leads to incorrect vowel predictions, severely degrading performance.

## 6. Conclusion

We have presented a framework for phonetically accurate Hebrew TTS, consisting of Phonikud for G2P conversion, the IL-Speech corpus with expert IPA annotations, a benchmark for Hebrew G2P, and audio-to-IPA models for TTS evaluation. Our experiments show that resolving phonetic underspecification enables small, local TTS models to approach the quality of large proprietary systems. Promising future directions include fine-grained prosody control, support for code-switching, and text normalization for symbols such as dates and addresses. More broadly, we see promise in adapting our insights to other languages facing related challenges due to phonetic underspecification, most directly for diacritized languages (Arabic, Urdu, etc.) and more generally for languages with opaque orthographies.

**Limitations.** Our method inherits limitations from the underlying Hebrew diacritization model, such as occasional vowel inaccuracies and adherence to formal written Hebrew conventions which may diverge from spoken norms (e.g. formal /sigrˈi/ vs. informal /sgerˈi/ for סגרי "close! (f.)"). These limitations are shared with existing Hebrew TTS methods.

## 7. Acknowledgements

We thank Dicta for encouraging our work and for approving our release of our model and data which incorporate their results. We also thank the speakers in ILSpeech for providing an essential resource for the development of Hebrew language technologies. We thank Oron Kam for assistance with ASR development. Finally, we acknowledge Kush Jain, Shlomo Tannor, Mark Kahn, and Shinji Watanabe for their helpful feedback and suggestions.

## 8. Generative AI Use Disclosure

Generative AI tools were only used for minor polishing of wording in this manuscript and coding assistance.